\documentclass[letterpaper, 10 pt, conference]{ieeeconf}  % Comment this line out if you need a4paper
\usepackage[dvipsnames]{xcolor}

\IEEEoverridecommandlockouts                              % This command is only needed if 
\usepackage{graphics} % for pdf, bitmapped graphics files
\usepackage{epsfig} % for postscript graphics files
\usepackage{mathptmx} % assumes new font selection scheme installed
\usepackage{times} % assumes new font selection scheme installed
\usepackage{amsmath} % assumes amsmath package installed
\usepackage{amssymb}  % assumes amsmath package installed
\usepackage{threeparttable}
\usepackage{graphicx}
\usepackage{subfig}
\usepackage{booktabs}
\usepackage{url}
\usepackage{hyperref}

\title{\LARGE \bf
Open Stamped Parts Dataset*
}

\author{Sarah Antiles$^{1}$ and Sachin Talathi$^{2}$% <-this % stops a space
\thanks{*This work was supported by General Motors}% <-this % stops a space
\thanks{$^{1}$Sarah Antiles, primary contributor
        {\tt\small sarah.antiles@gm.com}}%
\thanks{$^{2}$ Sachin Talathi, senior author
        {\tt\small sachin.talathi@gm.com}}%
}

\begin{document}

\maketitle
\thispagestyle{empty}
\pagestyle{empty}

%%%%%%%%%%%%%%%%%%%%%%%%%%%%%%%%%%%%%%%%%%%%%%%%%%%%%%%%%%%%%%%%%%%%%%%%%%%%%%%%
\begin{abstract}

We present the Open Stamped Parts Dataset (OSPD), featuring  synthetic and real images of stamped metal sheets for auto manufacturing. The real part images, captured from 7 cameras, consist of 7,980 unlabeled images and 1,680 labeled images. In addition, we have compiled a defect dataset by overlaying synthetically generated masks on 10\% of the holes.  The synthetic dataset replicates the real manufacturing environment in terms of lighting and part placement relative to the cameras. The synthetic data includes 7,980 training images, 1,680 validation images and 1,680 test images, each with bounding box and segmentation mask annotations around all holes. 10\% of the holes in the synthetic data mimic defects generated in the real image dataset. We trained a hole-detection model on the synthetic-OSPD, achieving a modified recall score of 67.2\% and a precision of 94.4\% . We anticipate researchers in auto manufacturing use OSPD to advance the state of the art in defect detection of stamped holes in the metal-sheet stamping process. The dataset is available for download at: \href{https://tinyurl.com/hm6xatd7}{https://tinyurl.com/hm6xatd7}.  

\end{abstract}

%%%%%%%%%%%%%%%%%%%%%%%%%%%%%%%%%%%%%%%%%%%%%%%%%%%%%%%%%%%%%%%%%%%%%%%%%%%%%%%%
\section{INTRODUCTION}

In auto manufacturing, sheet metal stamping is a cost-effective way to shape and cut flat metal sheets to produce several customized components of the vehicle at high-volume \cite{bobadestamp17}. 
Sheet metal stamping is a complex process where precise design and tooling is required to minimize defects and ensure optimal part quality. A multitude of defects, while rare, may still occur during production including but not limited to, splits \cite{singh2023HDR}, imprints \cite{block20inspection} and missing stamped holes. 

%Historically, sheet metal design heavily relied on trial-and-error methods and the expertise of skilled designers \cite{WAGENER1997342}.  Today, simulations and models offer assistance by predicting likely defect locations or by detecting defects \cite{MAKINOUCHI199619}. Our current efforts specifically target detection of missing hole defects. %Moreover, it is critical that issues during stamping be caught in a timely manner to limit damage downstream in the manufacturing process.

Manual inspection of stamped parts is time-consuming and susceptible to human errors. During this process, inspectors remove sheets from the production line to individually check for defects such as missing stamped holes. This is a laborious task due to the numerous, sometimes very small, holes. Consequently, manual inspection of every sheet is impractical. The high cost of manual inspection coupled with the costly consequence of defects, have spurred research interest in automated defect detection systems. Automation promises increased reliability, scalability, and cost-effectiveness.

In manufacturing, machine vision is increasingly used for part inspection. This involves capturing images via optical sensors followed by image processing to extract meaningful information. Processing techniques range from more traditional computer vision approaches to more advanced deep learning approaches. Because auto manufacturing settings are particularly dynamic, deep learning techniques, which learn pertinent information from the data, are advantageous \cite{singh2023HDR}. 

Deep learning spans various tasks including image classification, object detection, semantic segmentation, and instance segmentation. Our focus is on object detection because we require real-time localization and classification of defects, which we treat as individual objects.

The lack of publicly available training data poses a challenge for leveraging deep learning to develop models for sheet metal defect detection. Common object detection models such as YOLO \cite{yolo} and Faster R-CNN \cite{DBLP:journals/corr/RenHG015} are often trained on datasets like ImageNet \cite{5206848} and COCO \cite{lin2015microsoft}, which belong to the category of natural images and are distinctly different from images of sheet metal captured in an industrial setting. 

\begin{figure}
    \centering
    \subfloat[]{\includegraphics[width=0.48\linewidth]{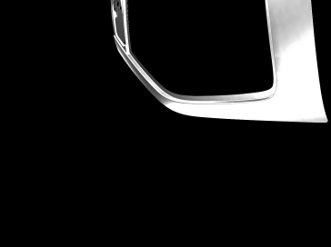}\label{fig:synthetic_view}}
    \hfil
    \subfloat[]{\includegraphics[width=0.48\linewidth]{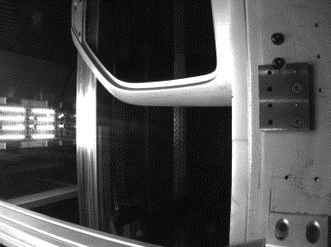}\label{fig:real_view}}
    \caption{Example of synthetically rendered (a.) and real stamped sheet metal (b.)}
    \label{fig:Fig1}
\end{figure}

Addressing the data challenge could involve training object detection models using synthetically generated images. Still, replicating the manufacturing environment proves difficult, often leading to a domain gap between synthetic and real data \cite{tremblay2018training}. In our dataset, one such gap lies in the background of the stamped sheet: synthetic images have no background artifacts, while real images depict conveyors and wire meshes. Refer to Fig. \ref{fig:Fig1} for a visual representation of this domain gap in OSPD. Recent strides in generative deep learning techniques offer promise in minimizing this gap, enhancing the robustness of object detection solutions \cite{wang2022parallelteacher, petsiuk2022synthetictoreal}.

% \begin{figure}[t]
%     \centering
%     %   \framebox{\parbox{3in}{\includegraphics[scale=0.5]{Fig1.pdf}}
%     % }
%     \includegraphics[width=1\columnwidth]{Fig1.pdf}

%     \caption{a. Example of synthetically rendered stamped metal sheet b. Example of real stamped sheet metal on conveyor belt in factory setting.}
%     \label{Fig1}
% \end{figure}

In order to facilitate these types of investigations, our work is focused on compiling a dataset of real sheet metal parts coupled with a large set of labeled synthetic sheet metal parts. We also develop a baseline hole detection model and present model performance numbers in terms of modified-recall and precision scores, specifically designed to evaluate the potential utility of the model when deployed on the factory floor.

Our hope is that the proposed baseline model, evaluation metrics, and datasets (both real and synthetic) of sheet metal images, the Open Stamped Parts Dataset (OSPD), will facilitate further development of efficient and reliable machine vision models to automate the process of defect inspection in auto manufacturing. Moreover, the provided datasets can serve as a playground for testing ideas which can then be generalized across different factory settings and manufactured parts.

Our key contributions are summarized as follows:
\begin{itemize}
    \item A labeled set of grayscale real images with bounding box annotations \footnotemark[1]$^{,}$\footnotemark[2]$^{,}$\footnotemark[3].
    \item A labeled set of grayscale real images with simulated missing hole defects and bounding box annotations \footnotemark[1]$^{,}$\footnotemark[2]$^{,}$\footnotemark[3]$^{,}$\footnotemark[4].
    \item A large scale, labeled synthetic dataset to mimic the real data with segmentation masks and bounding boxes\footnotemark[1]$^{,}$\footnotemark[3]$^{,}$\footnotemark[5].
    \item A large scale, unlabeled set of real images\footnotemark[1].
    \item Model benchmarks for hole detection on stamped metal parts using state-of-art deep learning techniques for real-time inspections.
    \item Modified recall metric for evaluating models based on desire to penalize defects resulting from stamping error.
    \footnotetext[1]{Metadata includes camera name and trigger distance.}
    \footnotetext[2]{See Fig. \ref{fig:entrance_and_profile} for camera fixture diagram. Trigger distance (measured in millimeters) is the distance the sheet metal travels along the conveyor relative to the fixture entrance.}
    \footnotetext[3] {Metadata includes whether the hole is graded. A graded hole has all hole edges visible in the reference image and is 100 pixels or larger. A reference image is a single real image for each camera and trigger distance combination (a ``view") without any sheet metal shift or camera shift relative to the expected positions.}
    \footnotetext[4]{Metadata includes whether or not the hole is masked.}
    \footnotetext[5]{Metadata includes the angle between the hole center's relative normal and the camera.}
    %for penalizing false positives only located near expected hole locations and penalizing all false negatives for graded holes \footnote{Graded holes are defined in metadata. Graded holes should be ones that have high contrast. Note, the graded holes are determined according to one inspection; a graded hole may not have high contrast in all inspections due to any number of production environment changes.}. \textcolor{red}{Do we need to explain this motivation in the above?}
\end{itemize}

\section{RELATED WORK}

\subsection{Inspection of Stamped Parts in Auto Manufacturing}
%Here, we summarize some of the recent published works that use deep learning methods for inspection of stamped parts in auto manufacturing.  

Singh \textit{et. al} introduce a YOLO variant trained on High Dynamic Range (HDR) photos to identify neck and split defects on sheet metal \cite{singh2023HDR}. Guo \textit{et. al} present an approach tailored for manufacturing, utilizing solely unlabeled real images to detect metal artifacts \cite{guo2021uir}. In our research, we incorporate unlabeled images for training, augmenting them with labels obtained from synthetic data. Block \textit{et. al} employ RetinaNet to detect metal part defects, later employing the MOSSE tracker algorithm to correlate detections over time, effectively mitigating false positives \cite{block20inspection}. False positives, which we define as the detection of a hole which does not exist, pose a significant concern in our context if the hole category deemed as found is in fact missing. Further details regarding our false positive reduction filtering method are elaborated on in the \textit{Model} Description section.

\subsection{Surface Inspection Datasets}
%Here, we summarize publicly available datasets for defect detection in the manufacturing space. 
In Huang \textit{et. al}, 1,386 images of circuit boards with 6 types of defects are labeled for detection, classification, and registration tasks \cite{huang2019pcb}. Real circuit boards are augmented with a synthetic defect generated with Adobe® Photoshop® software. One defect category, the ``missing hole", presents itself as a ring-like artifact with uniform green background. In contrast, the OSPD missing hole defect may not be distinguishable from the rest of the metal sheet. Thus, the focus in OSPD is to detect existing holes, which exhibit highly variable texture largely consisting of the background behind the stamped metal sheet. 

Other datasets include 18,074 images of flat steel sheets with four labeled defect types \cite{severstal}, 16,100 images of synthetically generated textured surfaces designed for industrial optical inspection with the defect labeled at the image level \cite{DAGM}, and 1,800 images of real steel surface inspections with 6 labeled defect types \cite{SONG2013858}. Unlike these datasets where the object of interest is on the metal surface itself, in OSPD, the object of interest is the stamped hole. A further comparison of these datasets to the OSPD can be found in Table \ref{tab1}.% Note that whereas all other surface inspection datasets focus on defects with detectable features, the missing hole defect in our dataset has no defining feature set.

\begin{table*}[t]
\caption{Public Surface Inspection Datasets}
\renewcommand{\arraystretch}{2}
\begin{center}
\begin{tabular}{|p{2cm}|p{2cm}|p{1.0cm}|p{3cm}|p{1cm}|p{6cm}|}
\hline 
\textbf{Dataset} & \textbf{Image Type}&  \textbf{\#Images}&  \textbf{Task} &\textbf{\parbox{4cm}{Synth. / \\Real}} & \textbf{Defect Classes}\\
\hline
PCB \cite{huang2019pcb}& HDR RGB&1,386&  \mbox{Classification}, \mbox{Detection}, \mbox{Registration} & Synth.  & open circuit, short, spurious copper,  missing hole, mouse bite, spur\\
\hline
Severstal \cite{severstal}  & grayscale & 18,074& \mbox{Semantic} \mbox{Segmentation} & Real &  \mbox{pitted}, \mbox{crazing}, \mbox{scratches}, \mbox{patches}\\
\hline
DAGM \cite{DAGM} &  grayscale & 16,100& Detection & Synth. & texture anomaly\\
\hline
NEU \cite{SONG2013858} &  grayscale & 1,800& Detection & Real &  rolled-in scale, patches, crazing, pitted surface, \mbox{inclusion}, scratches\\
\hline
OSPD &  grayscale& 11,340 & \mbox{Detection}, \mbox{Instance Segmentation} & Synth. & missing hole\\
\hline
OSPD &  \parbox{4cm}{grayscale} & 9,660& \parbox{4cm}{Detection} & Real & missing hole\\
\hline
\end{tabular}
\label{tab1}
\end{center}
\end{table*}

\subsection{Transfer Learning Between Synthetic and Real Data}
When a new part is onboarded, little to no real data is available for training a neural network. Training on synthetically generated data becomes imperative in these scenarios, and several works aim to bridge the gap between the real and synthetic data through different techniques \cite{wang2022parallelteacher}, \cite{petsiuk2022synthetictoreal}.  See the \textit{Experiments} section for a performance comparison of models trained on synthetic data that has and has not been transformed by an image-to-image (I2I) translation model.

\section{DATA COLLECTION PROCESS}

\subsection{Camera Fixture Setup}
\begin{figure}
    \centering
        \includegraphics[width=\linewidth]{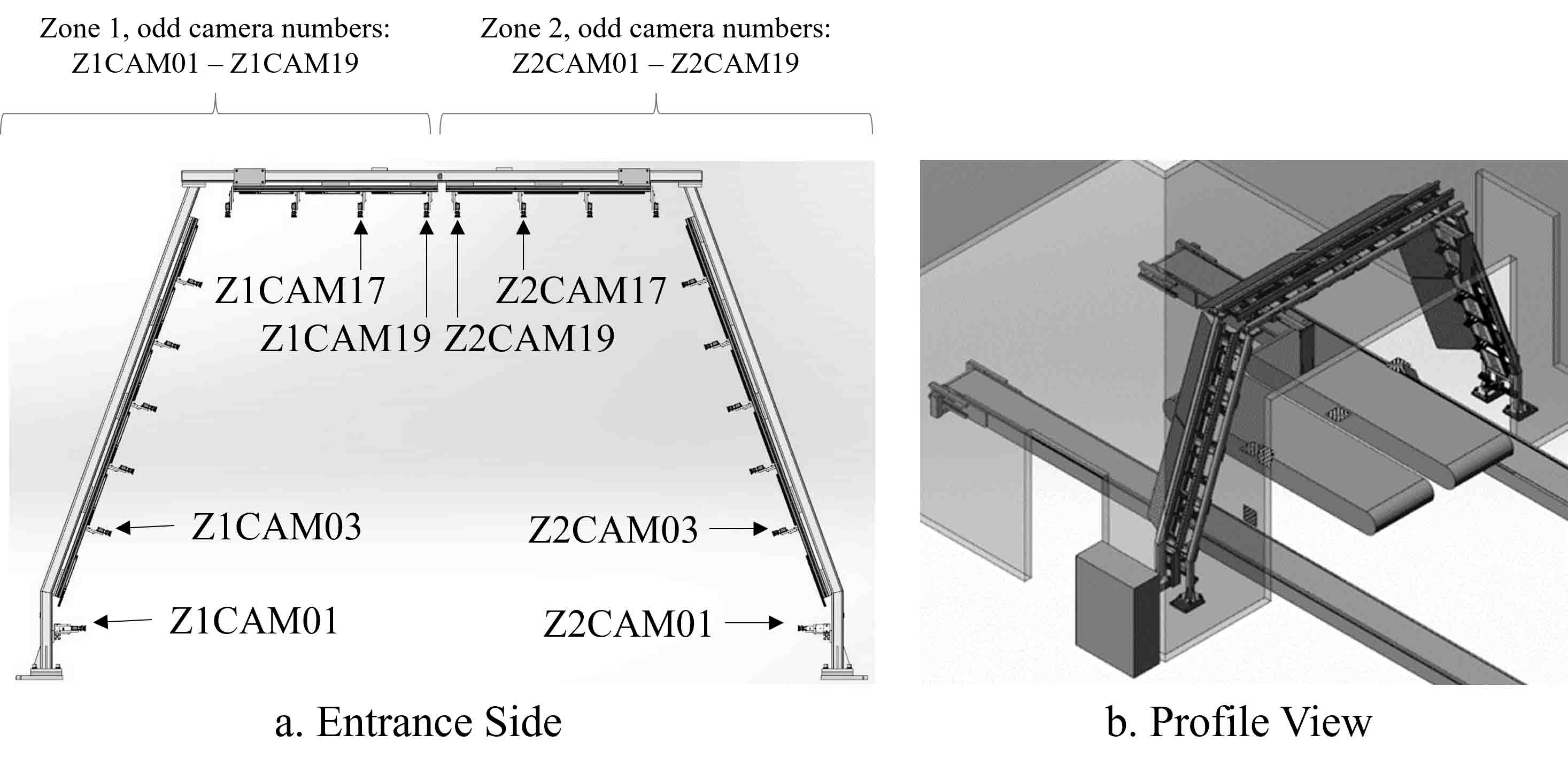}
        \caption{a. Entrance side fixture with subset of cameras labeled. Exit side mirrors entrance side, but with even-numbered cameras b. Illustration of conveyor passing through fixture.}
        \label{fig:entrance_and_profile}
\end{figure}

The OSPD's image collection uses a fixture with 40 cameras positioned across the left, right, entrance, and exit sides. 7 left-sided cameras took 21 unique captures in the OSPD. Entrance/exit designations are relative to conveyor belt movement, while right/left are relative to viewing from the entrance side. Cameras are named ``ZxCAMy," where x=1 and x=2 specify left and right-side camera respectively and \( y \in [1, 20] \). Refer to Fig. \ref{fig:entrance_and_profile} for fixture diagram with camera names. Each camera may trigger multiple times as the part moves through the fixture. OSPD metadata includes camera name (``cam\_name") and trigger distance (``trigger\_mm"). Lights on the fixture are infrared and cameras have infrared filters to block other light sources.

\subsection{Data Generation and Acquisition}
The synthetic data was acquired using Unreal Engine \cite{unrealengine}, by carefully configuring parameters such as camera angle, focal length, and lighting to replicate the actual production environment. Small amounts of random camera shift, part shift, and lighting changes were included to create synthetic data to simulate changes in production conditions. The simulation generated images of a stamped sheet with 7 distinct hole categories. To mimic missing stamped-hole defects, we applied a mask to a hole in any image with 10\% probability. We note that the conveyor and other background elements present in the manufacturing facility were not generated in Unreal Engine, resulting in a domain gap between synthetic and real data, which we aim to address.  

\begin{figure}
    \centering
    \subfloat[]{\includegraphics[width=0.48\linewidth]{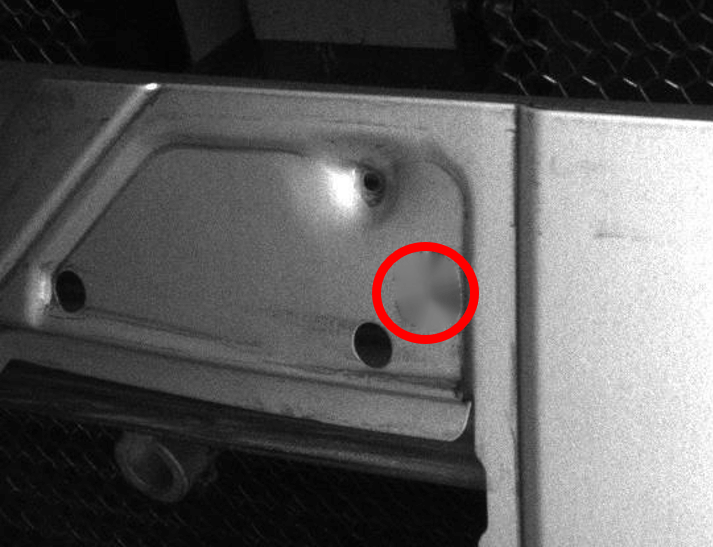}\label{fig:maskedimage1}}
    \hfil
    \subfloat[]{\includegraphics[width=0.48\linewidth]{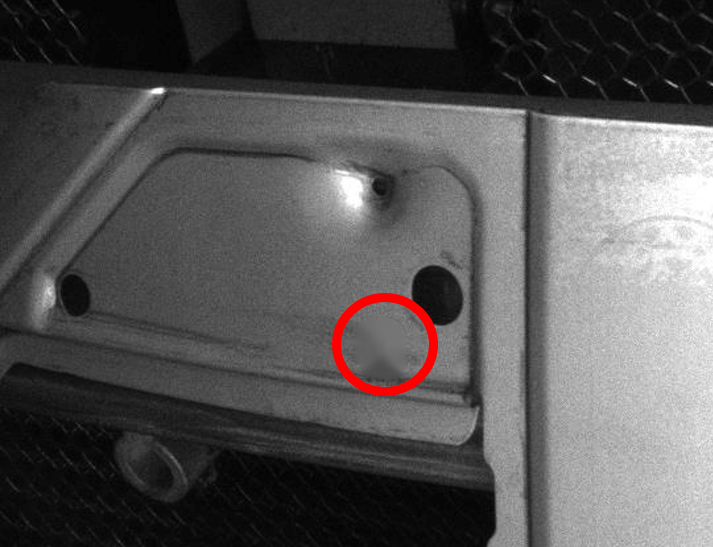}\label{fig:maskedimage2}}
    \caption{Examples of masked holes, hole category DY1 (a.) and hole category DY2 (b.)}
    \label{fig:masked_images}
\end{figure}

The real data used for testing was taken over the course of two days in March and April, 2023.  To assess a model's ability to identify missing holes accurately, while lacking actual instances of missing holes in real data, we transformed the test dataset by applying masks to a subset of the holes using an inpainting technique \cite{telea2004inpainting}. Refer to Fig. \ref{fig:masked_images} for representations of two different masked holes. 

% \begin{figure}
%     \centering
%     \includegraphics[width=1.0\linewidth]{masked_and_unmasked.png}
%     \caption{Examples of masked holes.  a. Hole category, DY1, outlined in red as masked b. Hole category, DY2, outlined in red as masked.}
%     \label{fig:masked_images}
% \end{figure}

We randomly assigned a mask to each hole in the real data with 10\% probability, resulting in 665 masked holes while the synthetic data has 4,964 masked holes. Because holes were masked at random, the percentages of masked holes in synthetic versus real data are comparable, though not identical.

Unlabeled real data was collected over 12 inspections days in August and September, 2023. Inspections were taken by the same cameras and trigger distances as the inspections in the labeled real data. Unlabeled real data is provided in the same quantity as the synthetic training set: 380 instances for each of the 21 views for a total of 7,980 images.

\subsection{Labeling Process}
Five individuals manually labeled the stamped holes using Label Studio \cite{Label}. To ensure accuracy and consistency, each person's labels underwent review by two others. Randomized labeling assignments prevented any individual from exclusively labeling all instances of a particular view.

\begin{figure}
    \centering
    \subfloat[]{\includegraphics[width=0.48\linewidth]{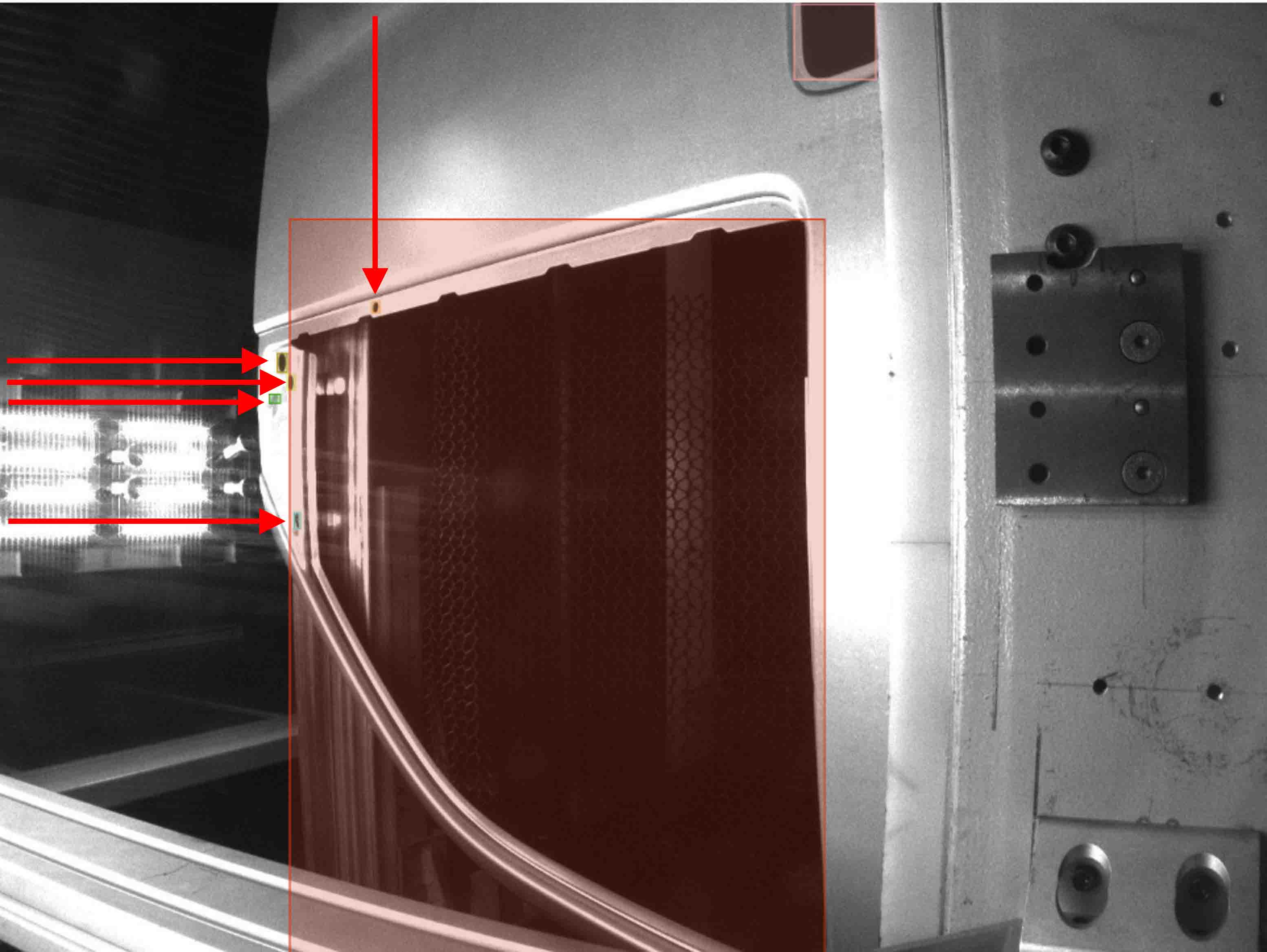}\label{fig:label_far}}
    \hfil
    \subfloat[]{\includegraphics[width=0.48\linewidth]{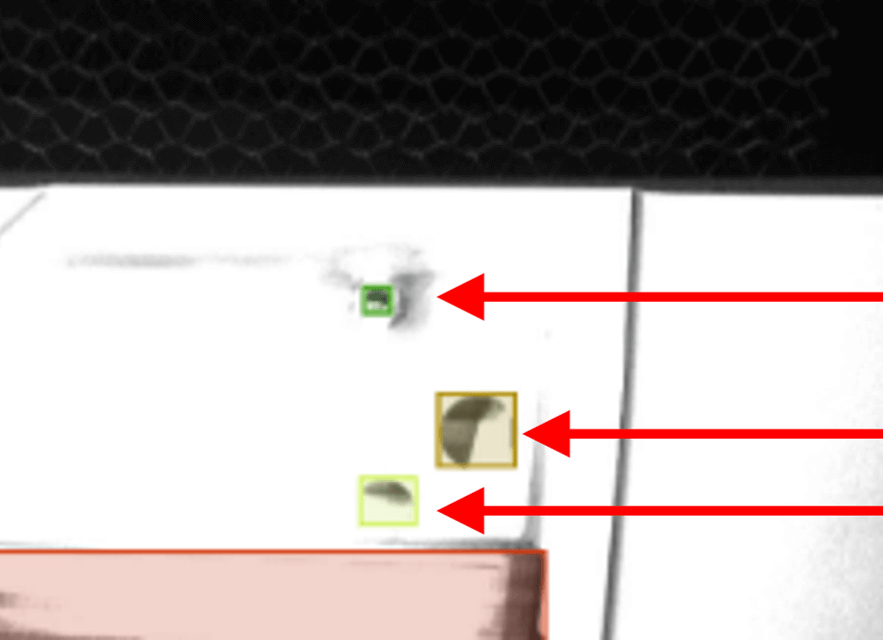}\label{fig:labe_contrast}}
    \caption{Examples of labeled hole images, small holes (a.) and partially visible, low contrast holes (b.) Arrows point to the referenced hole annotations.}
    \label{fig:labeled_images}
\end{figure}

Labelers were instructed to create bounding boxes around entire holes, despite potential difficulties in detection, given prior knowledge that all holes were present during inspections. Examples of challenging-to-detect holes, appropriately annotated, are depicted in Fig. \ref{fig:labeled_images}. For testing purposes, only graded holes were considered. See footnote 3 on page 2 for the definition of a graded hole. Additional details regarding evaluation metrics are available in the \textit{Metric} section. 

To create masked holes in the real data, we had a group of six labelers annotate holes with polygon masks in Label Studio. They were instructed to trace the hole's contour as accurately as they could. Subsequently, each label was reviewed by two other labelers for verification.

\section{IMAGE SETS \& STATISTICS}

We provide five datasets: 

\begin{enumerate}

\item{\bf Synthetic:} data generated by Unreal Engine; subset of holes masked; split into train, validation and test sets.
\item{\bf Real, labeled, reference:} available for use in training; only graded holes labeled; no masked holes.
\item{\bf Real, unlabeled:} available for use in training; no masked holes.  
\item{\bf Real, labeled:} dedicated for model testing only; no masked holes.
\item{\bf Real, labeled, 10\% masked:} dedicated for model testing only; is copy of ``real labeled" data with subset of holes masked.

\end{enumerate}

%All datasets are taken from the same 21 unique camera, trigger distance combinations which we call the ``view".  The synthetic dataset comprises 540 images from each of the 21 views, totaling 11,340 images. This synthetic dataset is further divided into training, validation, and test subsets, distributing the images per view as follows: 380 for training, 80 for validation, and 80 for testing. The real, labeled, reference images include 21 images, one per view. The real, unlabeled data mirrors the quantity of images in the synthetic training set for a total of 380 images per view for a total of 7,980 images. The real, labeled and real labeled, 10\% masked datasets each contain 80 images per view for a total of 1,680 images per dataset. 

\begin{figure}
    \centering
    \includegraphics[width=1\linewidth]{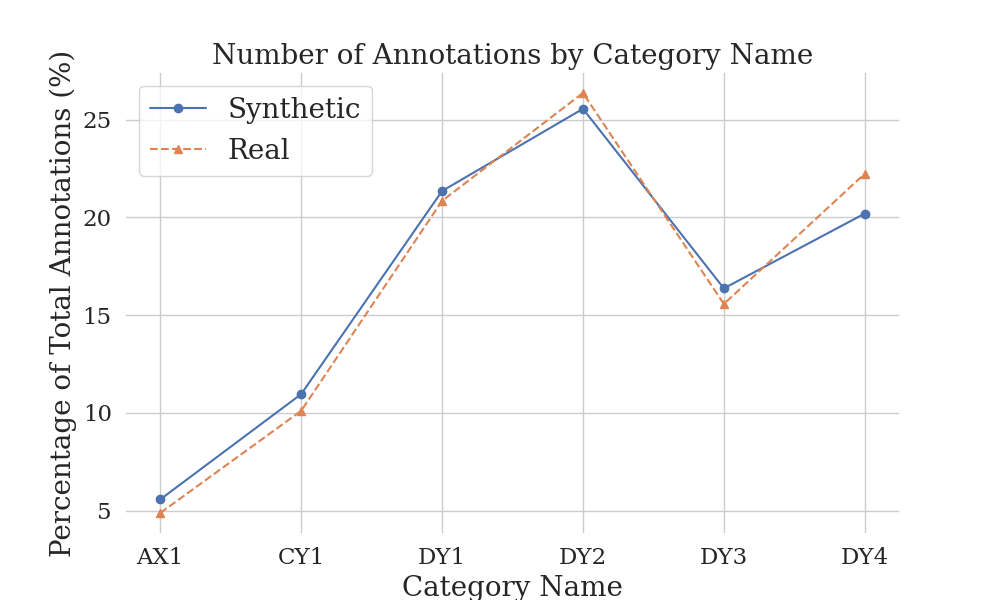}
    \caption{Number of graded annotations by category for synthetic and real data.}
    \label{fig:num_annotations}
\end{figure}

Fig. \ref{fig:num_annotations} shows the number of graded holes in the synthetic data and real data for each hole category. While absolute numbers for the synthetic data are larger than those for the real data, the relative quantities of each hole category across synthetic and real are comparable. 

Since missing holes have no distinctive features, we invert our problem: we formulate the task as detection of holes and thereby, we infer which holes are missing. In practical production scenarios, the occurrence of a missing hole is extremely rare (\textless1\% chance); however, the consequences of not detecting a missing hole are high.
 
% Considering that neural networks learn from examples, and noting that there are limited instances of missing holes in actual production data, we opted to increase the probability of missing holes in the synthetic training data beyond the expected frequency in real-world scenarios. \textcolor{red}{We recognize this could introduce bias in the model, however we hope to account for this in the penalty term of our modified recall metric.}

Since the recording session used to compile the real dataset didn't identify any missing holes, we have provided a separate dataset named ``real, labeled, 10\% masked", where masked holes are artificially generated. This dataset serves the purpose of evaluating a model's ability to properly recognize the absence of a hole.

\begin{figure}
    \centering
    \includegraphics[width=1\linewidth]{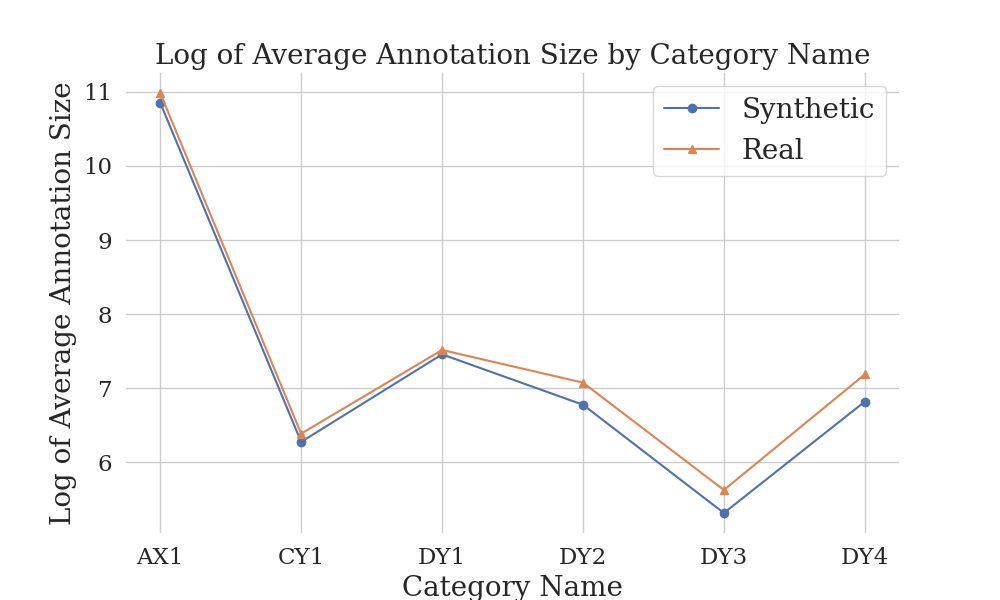}
    \caption{Average bounding box area in log scale by category for graded annotations in synthetic and real data.}
    \label{fig:area_combo}
\end{figure}

The average size (measured in pixels per inch) of each graded hole, which is calculated as the area of the bounding box, is shown in Fig. \ref{fig:area_combo}  for both synthetic and real, labeled data. On average, hole category AX1 is significantly larger than the other graded holes. Differences in the distribution of the area of each hole by category between synthetic and real is largely due to part shift relative to the camera, which is simulated in synthetic data and observed in real data. For further detail on number of holes by category and camera view, please refer to supplementary statistics at: \href{https://tinyurl.com/hm6xatd7}{https://tinyurl.com/hm6xatd7}.

% \begin{figure}
%     \newlength{\tempheight}
%     \setlength{\tempheight}{15ex}
%     \centering%
%     \begin{subfigure}[t]{0.5\textwidth}
%         \centering%
%         \includegraphics[totalheight = \tempheight]{annotation_area_synthetic_camname.png}
%         \caption{First picture.}
%     \end{subfigure}%
%     \begin{subfigure}[t]{0.5\textwidth}
%         \centering%
%         \includegraphics[totalheight = \tempheight]{annotation_area_synthetic_trigger.png}
%         \caption{Second picture with a longer caption that will need more lines than the one from the first picture.}
%         \label{fig:sub2}
%     \end{subfigure}
% \end{figure}

% \begin{figure}
%     \centering
%     \includegraphics[width=1\linewidth]{scatter_bbox_area_size_real.png}
%     \caption{Average bounding box area in log scale by camera, trigger distance, and category for real, labeled, unmasked data. Note, only graded holes are considered.}
%     \label{fig:area_real}
% \end{figure}

\begin{figure}
    \centering
    \includegraphics[width=1\linewidth]{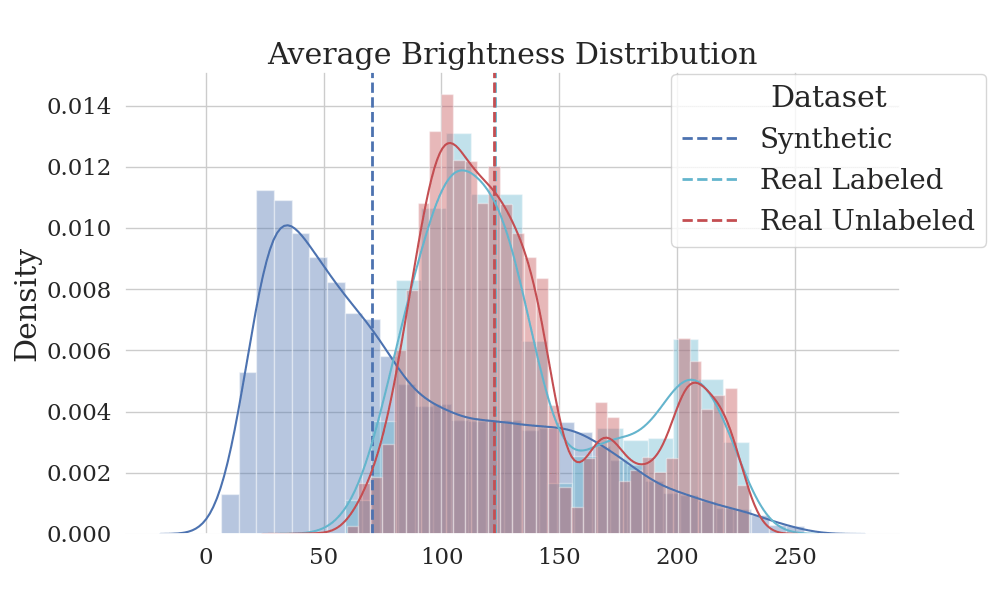}
    \caption{Average image brightness distributions across datasets. Vertical dotted lines indicate distribution medians.}
    \label{fig:brightness_distribution}
\end{figure}

 Fig. \ref{fig:brightness_distribution} compares the brightness levels across three datasets: synthetic, real labeled, and real unlabeled. Notably, relative to the brightness distribution of synthetic data, the brightness distributions of the real unlabeled and real labeled images show a stronger resemblance. 

%The median brightness of the synthetic I2I images appears to align more closely with the real data than that of the synthetic (original) dataset. This suggests that employing I2I data may potentially enhance model performance by bringing the synthetic data closer to the characteristics of real data. 

% We measure the texture and pixel distribution to understand how similar the synthetic data is to the real data. Figure \ref{fig:brightness_distribution} compares brightness across the synthetic (original) images, synthetic I2I images, real unlabeled images and real labeled images. The real unlabeled and real labeled images come from relatively similar distributions compared with the synthetic data. The synthetic I2I median brightness is closer to the real data than that of the synthetic (original) suggesting I2I data may improve model performance. We also compute Frechet Inception Distances (FID) between synthetic (original), synthetic (I2I) and real unlabeled datasets shown in figure \ref{fig:fid_dist} \cite{heusel2018gans} \textcolor{red}{should i take fid table out since does not show synthetic I2I closer to real than synthetic original}. FID distances were calculated with a subset of 5,000 images from each dataset. While exact FID values are not meaningful, it is evident that I2I techniques provide a path towards bridging the gap between synthetic and real images, though there is still a meaningful distance between the synthetic (I2I) data and the real data. 

% \begin{figure}
%     \centering
%     \includegraphics[width=.5\linewidth]{FID_Distance.png}
%     \caption{FID distances between synthetic and real data. 5000 images are sampled from each dataset for the calculation.}
%     \label{fig:fid_dist}
% \end{figure}
We note that masked holes are undersampled in the data relative to unmasked holes. It is however crucial to highlight that the shape and texture of a missing hole are not well-defined in practical production scenarios. This factor motivates our emphasis on detecting holes rather than specifically targeting the detection of missing holes. For a deeper understanding of how we address this undersampling during the evaluation of our models, please refer to our explanation of modified recall score in the \textit{Metric} section.

Not all hole categories are equally represented across the dataset or across different views. For instance, AX1, the largest graded hole, constitutes only 6\% of graded holes in the synthetic data and 5\% of graded holes in the real data. Hole BY1, though present, is not graded.

\section{MODEL DESCRIPTION}

\subsection{Object Detection}
YOLO, a leading CNN based object detection model serves as our baseline model \cite{yolo}.  Specifically, we train a YOLOv7 model\footnotemark[1]\footnotetext[1]{Code available at \url{https:// github.com/WongKinYiu/yolov7}}, as it demonstrates superior speed and accuracy compared to prior YOLO architectures \cite{wang2022yolov7}. The YOLOv7 model was trained for 1,000 epochs with a batch size of 8 on an NVIDIA A100 GPU. Training images were resized to 640x640, padding the smaller dimension of the original image to maintain the aspect ratio.

For training of the binary hole/no-hole models, we chose to assign categories to predictions through a post-processing step that rests on the assumption that each view has a defined set of expected hole categories given by the view's reference image. The procedure for filtering predictions and assigning category is as follows:

For each image, \(I\), in the test set, grab its corresponding reference image, \(R\), and for each annotation, \(f_{r}\), in the reference image, map it to a subset of predictions, \(\{i_{p}| i_{p} \subseteq I_{p}\}\) , where \(I_{p}\) is the set of model predictions on image, \(I\). We create this mapping by first considering all predictions as potential matches to the reference annotation and then filtering out predictions through a method as follows:
\begin{enumerate}
    \item If the area of \(f_{r}\), \(a_{r}\), is greater than 1,100 pixels, calculate the structural similarity measure (SSIM) \cite{1284395}, average local binary patch (LBP) \cite{liu2017local} and template match score (TMS) \cite{lewis2001FNCC} between the cropped image contained in \(i_{p}\) and \(f_{r}\) to obtain: \(SSIM_{i,f}\), \(LBP_{i,f}\) and \(TMS_{i,f}\). Filter out \(i_{p}\) where \(SSIM_{i,f}< 0.19\), Keep all \(i_{p}\) where \(SSIM_{i,f}> 0.45\), and for \(i_{p}\) where \(SSIM_{i,f}>0.19\) and \(SSIM_{i,f}<0.45\), do\footnotemark[2]\footnotetext[2]{Bounds derived empirically}:
    \begin{itemize}
        \item Filter out \(i_{p}\) where \(LBP_{i,f}>0.007\)
        \item Filter out \(i_{p}\) where the \(TMS_{i,f}<=0.925\)
    \end{itemize}
    If \(a_{r}<=1,100\), filter out all \(i_{p}\) except for \(i_{p}\) whose center is nearest to \(f_{r}\).\footnotemark[3]\footnotetext[3]{Euclidean distance is used for calculating distance \cite{euclideanprecalculus}}.
    After all \(f_{r}\) have been considered, each \(f_{r}\) will be assigned to zero, one, or many \(i_{p}\). 
    \item Find the maximum cardinality matching by solving the minimum cost assignment algorithm where cost values in our cost matrix are defined as the Euclidean distance between the center of \(i_{p}\) and \(f_{r}\) and the boolean matrix is derived from the previous step where we found potential matches. In this way, each \(i_{p}\) is assigned to at most one \(f_{r}\) and each \(f_{r}\) is assigned to at most one \(i_{p}\) \cite{Ramshaw2012OnMA}.
    \item Assign category ID to \(i_{p}\) according to the ID of its assigned \(f_{r}\).
\end{enumerate}
Note, step 3 is only used in binary-class models. For multi-class YOLO models, we perform steps 1 and 2 only.  

\subsection{Unpaired Image-to-Image Translation}

Image-to-Image (I2I) translation is a specific use case in generative modeling of digital images, where the objective is to synthesize an output image to match the texture and pixel distribution of images in the target domain from images in the source domain, while preserving the structure and content of the input image \cite{9528943,sym12101705}. In the context of sheet metal stamping, synthetic images belong to the source domain and the sheet metal images obtained from the factory floor belong to the target domain.

In our I2I experiments, we train a contrastive unpaired translation (CUT) image-to-image translation model using unpaired synthetic and real images \cite{park2020contrastive}. We then feed our synthetic data through this model to realify the data and train a YOLOv7 model. % The CUT model was chosen for its relative simplicity and low memory usage compared to diffusion models \cite{parmar2024onestep}.

\subsection{Metric}
Our foremost goal is to accurately identify missing holes; classifying a missing hole as found can significantly compromise this objective. Misclassification can occur in two ways: the model either recognizes a missing hole as a hole or identifies a different object as a hole. Consequently, our metric aims to reflect our strong preference for zero false positives in both of these scenarios. Provided there is a low occurrence of false positives on objects, our secondary objective involves minimizing instances where the model incorrectly labels an existing hole as missing. As such, we define a best model as one with the highest F$_{\beta}$ score: \begin{equation}
    F_\beta = (1 + \beta^2) \cdot \frac{{\text{precision} \cdot \text{recall}}}{{\beta^2 \cdot \text{precision} + \text{recall}}} 
    \end{equation} where $\beta=0.9$, showing our preference for models with higher precision over higher average modified recall. We define precision in the standard way\footnotemark[4].
``Modified recall" combines the traditional recall calculated on the real, labeled dataset with an additional penalty term for any predictions made on masked holes within the real, labeled 10\% masked dataset.
The equation for modified recall is as follows: \begin{equation}
  \frac{1}{J}\sum_{j=1}^{J}
r_{u,j} * \frac{e^{-1} fp_{m,j} +t_{m,j} - fp_{m,j}}{t_{m,j}}  
\end{equation}
Here, $J$ represents the total number of graded hole categories, $r_{u,j}$ represents the recall calculated on the real, labeled dataset of graded holes for hole category j; $fp_{m,j}$ represents the false positive predictions of hole category j, made on masked holes (graded or ungraded) within the real, labeled, 10\% masked hole dataset; and $t_{m,j}$ denotes the total number of masked holes in the real, labeled 10\% masked hole dataset\footnotemark[4].
%\textcolor{red}{In Table \ref{model results} we compare average recall with average modified recall. Without an extra penalty for detected masked holes, the F-Score for the I2I binary model exceeds that of the synthetic binary model, but given that there are several masked holes which are predicted as holes by the I2I model, we penalize this model more harshly such that the modified F-Score of I2I drops below the modified F-Score for the synthetic binary model.}

\footnotetext[4]{Intersection over union threshold is 0.2 if area of ground truth annotation$<96^{2}=9,216$, else, threshold is 0.5.}

\begin{table*}[ht]
\caption{Model Results}
\begin{center}
\begin{tabular}{|p{6cm}|p{2.0cm}|p{1.0cm}|p{1cm}|p{1.1cm}|p{1.cm}|p{1.5cm}|}
\hline 
\textbf{Model} & \textbf{Conf. Thresh.} & \textbf{AR}& \textbf{AMR}&  \textbf{AP} & \textbf{F$_{0.9}$ Score} & \textbf{Modified F$_{0.9}$ Score} \\ 
\hline
\mbox{Synthetic –} multi-class                          & 60 & 67.2 & 67.2 & 94.4 & 79.9 & 79.9\\
\hline
\mbox{I2I} – multi-class                               & 65 & 61.6 & 58.7 & 73.1 & 67.5 & 65.9\\
\hline
\mbox{Synthetic – binary + custom category ID protocol} & 75 & 61.1 & 61.1 & 81.6 & 61.1 & 70.9 \\
\hline
I2I – \mbox{binary + custom} \mbox{category ID} protocol       & 65& 60.3 & 57.7 & 86.4 & 72.4 & 70.7\\
\hline
\end{tabular}
%\caption*{\textcolor{red}{Note: Models trained on synthetic data and tested on the real, labeled and real, labeled masked test sets.}}

%The table illustrates the trade off between precision and modified recall. The models that output category ID directly favor modified recall at the expense of lower precision while the models that include the separate post-processing step, which assigns category ID to each prediction, favor precision at the expense of lower modified recall.
\label{model results}
\end{center}
\end{table*}

\section{EXPERIMENTS}
Table \ref{model results} displays the confidence threshold, average recall (AR), average modified recall (AMR), and average precision (AP) that yields the highest Modfied F$_{0.9}$ score where confidence threshold was tested at \(5\%\) increments from \(50\%\) to \(95\%\). Synthetic models are YOLOv7 models trained on data generated directly by Unreal Engine while I2I models are YOLOv7 models trained on data generated by Unreal Engine and then fed through a CUT image-to-image translation model \cite{park2020contrastive}.  Multi-class models predict each object as one of several hole classes and are fed through the filtering method outlined in the {\it Model Description} section for increasing precision. The binary + custom category ID protocol models predict each object under a single hole class and are fed through the protocol outlined in the {\it Model Description} section for filtering and category assignment.
% \begin{enumerate}
%     \item Synthetic – multi-class: The YOLO model is trained to predict each object as one of several hole classes. Training data is data generated directly by Unreal Engine.
%     \item I2I – multi-class: The YOLO model is trained to predict each object as one of several hole classes. Training data is data generated by Unreal Engine and fed through a CUT image-to-image translation model.
%     \item Synthetic – binary + custom category ID protocol: The YOLO model is trained to predict each object under a single hole class. Training data is data generated directly by Unreal Engine.
%     \item I2I – binary + custom category ID protocol: The YOLO model is trained to predict each object under a single hole class. Training data is data generated by Unreal Engine and fed through a CUT image-to-image translation model.
% \end{enumerate}
Because we are interested in optimizing the modified recall metric, we chose to evaluate each model using the model checkpoint which yields the highest recall on the validation set during training. Training a multi-class model with purely synthetic data yields the highest modified recall, 67.2, and highest precision, 94.4. I2I appears to add no increase in performance. One reason for this may be because we approached I2I in a simplistic way: splitting image into 16x16 patches, passing through I2I model, and stitching patches back together.

\section{FUTURE RESEARCH}
In future work one could expand the synthetic and real dataset to include more types of stamping defects.  One could investigate more advanced I2I techniques or background subtraction models to remove background distractions present in real data, bridging the domain gap between synthetic and real data. This may also improve the generalization capabilities of the defect detection model across different factory settings.

% \section{CONCLUSION}
% We introduced OSPD, a dataset comprising synthetic and real stamped metal sheets, which aims to enhance defect detection of stamped holes in manufacturing. Our custom metric prioritizes identifying missing holes, and we presented a baseline model as a starting point for future enhancements. Finally, our statistical analysis of the data offers contextual insights and identifies potential sources of model bias.    

\addtolength{\textheight}{-12cm}   % This command serves to balance the column lengths
                                  % on the last page of the document manually. It shortens
                                  % the textheight of the last page by a suitable amount.
                                  % This command does not take effect until the next page
                                  % so it should come on the page before the last. Make
                                  % sure that you do not shorten the textheight too much.
%%%%%%%%%%%%%%%%%%%%%%%%%%%%%%%%%%%%%%%%%%%%%%%%%%%%%%%%%%%%%%%%%%%%%%%%%%%%%%%%

%%%%%%%%%%%%%%%%%%%%%%%%%%%%%%%%%%%%%%%%%%%%%%%%%%%%%%%%%%%%%%%%%%%%%%%%%%%%%%%%

%%%%%%%%%%%%%%%%%%%%%%%%%%%%%%%%%%%%%%%%%%%%%%%%%%%%%%%%%%%%%%%%%%%%%%%%%%%%%%%%
% \section*{APPENDIX}

% Appendixes should appear before the acknowledgment.

\section*{ACKNOWLEDGMENT}
Thank you to Jose Armando Lopez  Jr. and Thomas Sandrisser for their support of this project. We also acknowledge support from Rosemary Binny, Sherry Courington, Carlos Cardenas, Beatrice Lovely, Anishi Mehta, Joseph Muniz Jr., Jose Rangel, Hilton Shumway, Sean Sloan and Anthony Swink, Tyler Wingo  and Steve Rangos. 

\bibliographystyle{ieeetr}
\bibliography{bibliography}

\end{document}